\documentclass[letterpaper, 10 pt, conference]{ieeeconf}  

\IEEEoverridecommandlockouts                              

\usepackage{epsfig} 
\usepackage{soul,color,graphicx,float,epstopdf}
\usepackage{tablefootnote,textcomp,gensymb}
\usepackage{eurosym,booktabs,multirow}
\usepackage{tabularx}
\usepackage{url}
\usepackage{times} 
\usepackage{amsmath} 
\usepackage[caption=false]{subfig}
\usepackage{cite}
\usepackage{xcolor}
\usepackage{amsfonts} 
\usepackage[hidelinks]{hyperref}
\usepackage{svg}
\newcommand{\boldpar}[1]{\smallbreak\noindent\textbf{#1}}
\DeclareSymbolFont{Xlargesymbols}{OMX}{cmex}{m}{n}
\DeclareMathSymbol{\Xsum}{\mathop}{Xlargesymbols}{80}

\title{\LARGE \bf
Modeling and Analysis of Multi-Line Orders in Multi-Tote Storage and Retrieval Autonomous Mobile Robot Systems
}

\author{Xiaotao Shan, Yichao Jin, Peizheng Li, Koichi Kondo
\thanks{X. Shan, Y. Jin, P. Li are with Bristol Research and Innovation Laboratory, Toshiba Europe Ltd., U.K. Emails: \{firstname.lastname\}@toshiba-bril.com.}
\thanks{K. Kondo is with Toshiba Corporation, Corporate Technology Planning Div., Kawasaki, Japan, Email: koichi1.kondo@toshiba.co.jp}}%

\begin{document}

\maketitle
\thispagestyle{empty}
\pagestyle{empty}

\begin{abstract}

As warehouses are emphasizing space utilization and the ability to handle multi-line orders, multi-tote storage and retrieval (MTSR) autonomous mobile robot systems, where robots directly retrieve totes from high shelves, are becoming increasingly popular. This paper presents a novel shared-token, multi-class, semi-open queueing network model to account for multi-line orders with general distribution forms in MTSR systems. The numerical results obtained from solving the SOQN model are validated against discrete-event simulation, with most key performance metrics demonstrating high accuracy. In our experimental setting, results indicate a 12.5\% reduction in the minimum number of robots needed to satisfy a specific order arrival rate using the closest retrieval sequence policy compared with the random policy. Increasing the number of tote buffer positions on a robot can greatly reduce the number of robots required in the warehouse.

\end{abstract}

\section{INTRODUCTION}
The exponential growth observed among e-commerce companies, coupled with rising labor costs and the necessity for careful handling of goods, has spurred a demand for Robotic Mobile Fulfillment Systems (RMFS)~\cite{teck2023efficient,justkowiak2023stronger,lu2023automated, shan2024distributed, wu2024proof}. In the context of a typical RMFS, robots directly transport shelves, referred to as 'pods', containing products between workstations and the shelf storage area~\cite{lamballais2017estimating}. However, directly transporting shelves imposes limitations on shelf height and weight, leading to reduced space utilization. Moreover, if an order requires multiple product lines, robots need to transport numerous shelves between the central storage area and workstations in several trips, thereby elongating robot travel times and diminishing system efficiency. The MTSR system is designed to address these challenges~\cite{hairoboticsHaiPickSystem, qin2024performance}, which allows robots to directly retrieve totes from shelves and transport multiple totes simultaneously within one trip~\cite{hairoboticsHaiPickSystem}. 

\begin{figure}[!ht]
    \centering
    \includegraphics[width=0.48\textwidth]{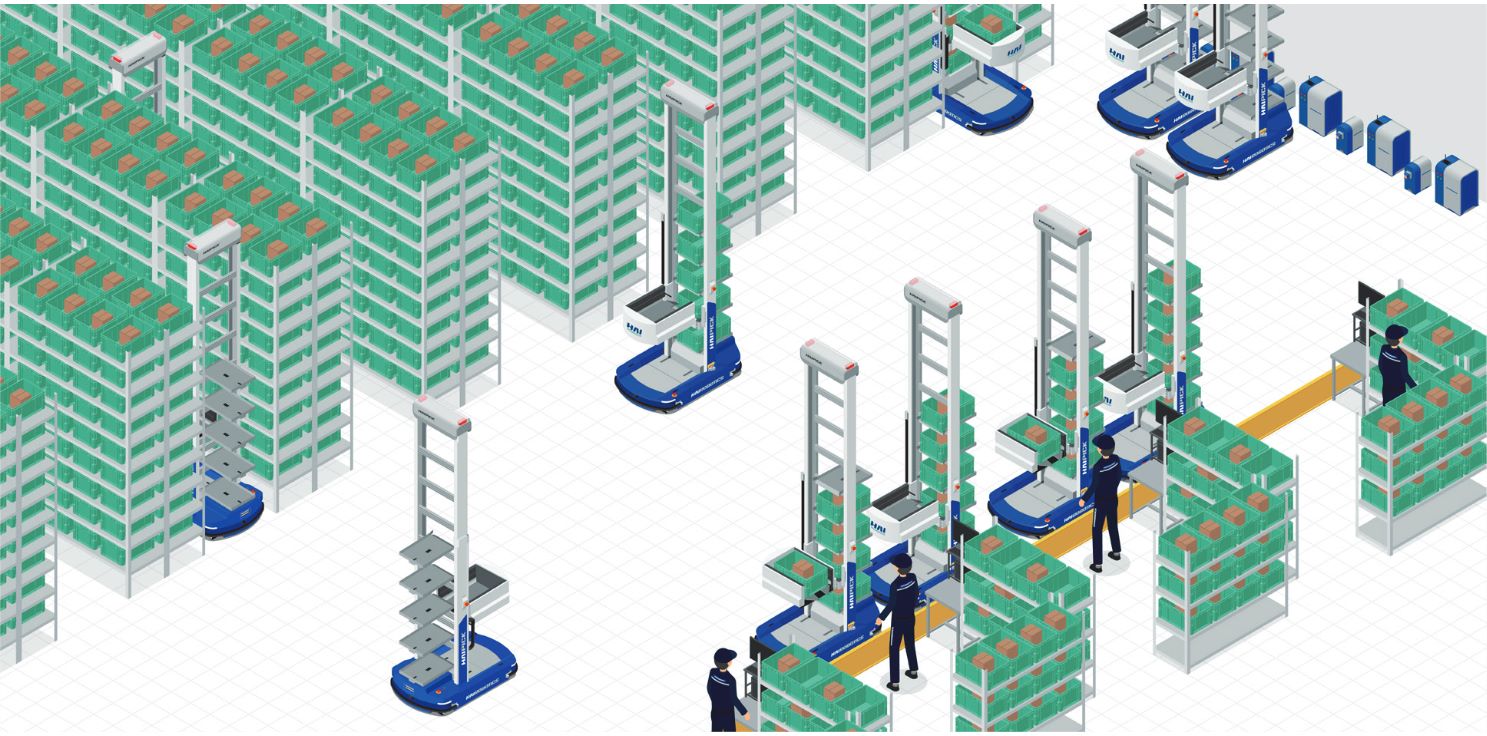}
    \vspace{-5pt}
    \caption{The MTSR autonomous mobile robot system, where robots directly pick multiple totes from vertical storage shelves~\cite{hairoboticsHaiPickSystem}.}
    \label{fig:ACRS}
    \vspace{-15pt}
\end{figure}
Queueing Network (QN) offers an approach to understanding complex logistics systems by representing various stages of the operational process as a network of interconnected nodes. The movement of robots within this network resembles state changes in a Markov chain. Solving this chain reveals the steady-state performance and resource bottlenecks of the logistics system more efficiently compared to simulation-based methods. QN analysis has been widely applied in 2D-RMFS~\cite{chi2021analysis, wang2022performance, yang2022modelling, wu2020research, zou2018evaluating, roy2019robot, lamballais2020inventory}. However, there is little research investigating the MTSR system, despite its increasing popularity across diverse fields, including retail, electronics, and healthcare~\cite{hairoboticsHaiPickSystem}.

Although the research in~\cite{qin2024performance} analyzes the MTSR system through a SOQN, it does not investigate the influence of orders with varying numbers of product lines on the system performance. This imposes some new challenges: \textbf{1)} The probability density function of the number of lines within an order can follow a general distribution, rather than being limited to only the geometric distribution~\cite{duan2021performance,lamballais2017estimating}; \textbf{2)} Due to the limited buffer positions on a robot, the robot might need to make several trips to fulfill an order if the number of lines within it exceeds the buffer positions; \textbf{3)} The service time at the workstation where goods are handled and the traveling time while retrieving and storing totes should depend on the number of assigned totes to the robots in a trip. Thus, our contributions in this paper can be summarized as follows:

\begin{itemize}
    \item We construct a novel shared-token, multi-class SOQN based on the operational process of the MTSR system to account for multi-line orders with general distribution. The steady-state performance, such as order throughput time and utilization of different resources, is obtained by solving the model using the approximate mean value analysis (AMVA).
    \item The numerical solutions are validated through discrete-event simulations. We investigate the impact of the robot's tote buffer positions, the number of robots, and different tote retrieval sequence policies (closest retrieval and random) on the system's performance. 
\end{itemize}

The remaining part of this paper is constructed as follows: Sec.~\ref{sec:literature review} demonstrates related works in the field of RMFS performance estimation. The system operational process and assumptions are detailed in Sec.~\ref{sec: System description}. The proposed SOQN and solution for the system is given in Sec.~\ref{sec: SOQN}. Sec.~\ref{sec: numerical experiments} describes the scenario setup and analyzes the experimental results, and Sec.~\ref{sec: conclusion} concludes this paper.

\section{LITERATURE REVIEW} \label{sec:literature review}
Study~\cite{qin2024performance} is the first paper to model the MTSR system. A mixed storage and retrieval policy (MSR) is discussed in the paper. In this pattern, the tote needed to be stored will be stored in the place that has just been retrieved. Although the original tote storage positions will be disturbed, this mode can save a lot of traveling time. Since the number of totes required to be stored and retrieved is large, the sequence of storing and retrieving totes becomes important. The closest retrieval sequence policy based on the nearest neighbor heuristic is investigated using a numerical method. Finally, an SOQN model is built to analyze tote throughput time and throughput capacity. In their model, it's assumed that in every trip, the orders assigned to robots have a fixed number of totes equal to the robot's buffer positions. However, in reality, the number of lines within an order varies, and the buffer positions of robots are not always fully occupied.



There are studies investigating multiple types of arriving orders, such as retrieval and replenishment orders, in RMFS. Zou \textit{et al}.~\cite{zou2018evaluating} assume that retrieval and storage orders share the same operational process. The originally constructed multi-class SOQN is simplified to a single-class SOQN for resolution. A multi-class closed queueing network (CQN) model is built in research~\cite{roy2019robot} to account for both retrieval and replenishment orders. They evaluate and compare the order throughput time for both dedicated and pooled robot systems for pod retrieval and replenishment. However, this model assumes that the robots are continuously busy, without considering the order arrival rate. Differing from these two models, Lamballais \textit{et al}.~\cite{lamballais2020inventory} correlate retrieval and replenishment processes by classifying orders and pods based on the required or contained number of units per stock keeping unit (SKU). Pods are replenished when the unit count of an SKU falls below a certain threshold and are used for orders if the count exceeds the required quantity. They construct a cross-class matching multi-class SOQN and analyze it using a Continuous Time Markov Chain (CTMC). However, the state count of the Markov chain escalates rapidly with the pod capacity, the number of pods distributed per SKU, and the types of SKU. Moreover, robot utilization is not considered in their analysis.

Multi-line orders in RMFS are studied in~\cite{lamballais2017estimating} and~\cite{duan2021performance}. These orders are presumed to follow a geometric distribution in terms of the number of lines. A multi-line order requires the same robot to pick from different shelves across multiple retrieval and storage trips. The average travel time for retrieving and storing a shelf, as well as the processing time distribution, remains consistent across trips. Thus, the single-class QN can handle multi-line orders by treating them as a bundle of identical single-line orders. However, the probability distribution of the number of lines within an order can vary in practice. Furthermore, the travel time for tote retrieval and storage, along with the robots' service time, may significantly vary across trips due to the different number of handled totes in MTSR system. Therefore, orders with varying line counts in MTSR system cannot be simply decomposed into a variable number of identical single-line orders, unlike the studies in~\cite{duan2021performance,lamballais2017estimating}.

\section{SYSTEM DESCRIPTION} \label{sec: System description}
\begin{figure}[htbp]
    \centering
    \includegraphics[width=\linewidth]{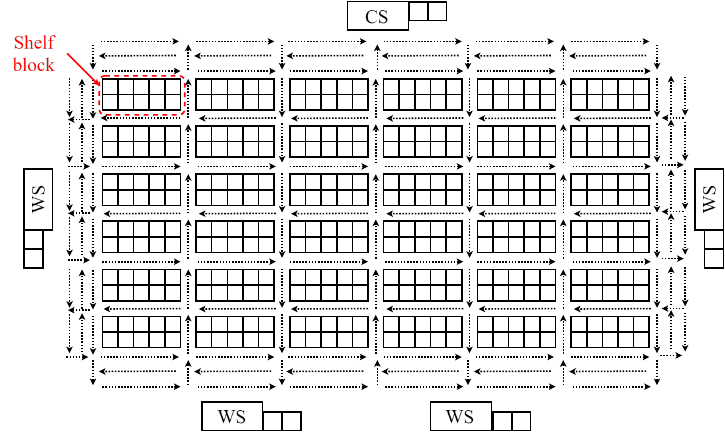}
    \caption{Top view of an MTSR system layout, where WS symbolizes the workstation for retrieving goods from arriving totes or replenishing them with goods, CS denotes charging station.}
    \label{fig:layout}
\end{figure}

Fig.~\ref{fig:layout} shows a classical layout of the warehouse, featuring a central shelf storage and retrieval area surrounded by work and charging stations. The workstation is designed to pick the required goods from the totes delivered by the robots or to replenish the totes with goods. Unidirectional aisles and cross aisles connect these areas, mitigating collisions and deadlocks.

The operation process of robots within the MTSR system can be summarized as follows:
\begin{enumerate}
\item Multi-line orders arrive and are assigned to an idle robot. In the MTSR system context, each line corresponds to a product stored in different totes. Robots have a limited tote buffer, necessitating multiple trips to fulfill an order. 
\item The MSR policy~\cite{qin2024performance} is not used in this paper because it makes the original storage position of the totes disrupted. The placement of the totes in many warehouses is optimized based on product relevance, which can help reduce the robots' traveling distance. Therefore, in this paper, the totes will be stored back in their original locations, and the tote storage and retrieval processes will be separated.
\item Robots plan a tote retrieval sequence and move from their current locations to retrieve totes required by the orders from shelves according to the planned sequence. In this paper, the closest retrieval (CR) sequence policy~\cite{qin2024performance}, based on the nearest neighbor heuristic, and the random retrieval sequence policy are investigated. For the nearest neighbor heuristic, the distance metric refers to the shortest distance from one point to another following the predefined unidirectional path network.
\item After the robot collects all the totes from the shelves, it transports them to the designated workstation.
\item The number of workers at the workstation is limited. A worker can only serve one robot at a time. If there is no idle worker, the robot enters the workstation buffer and waits for its turn.
\item Once the workers finish their operations on the totes, the robot returns the totes to their original locations in the storage area. The sequence for storing the totes could be either random or CR policy.
\item The robot then checks if there are any remaining totes needed for the order. If so, it repeats the process starting from step 2.
\item If no totes remain, the order is complete. Robots with low battery levels (below a predefined threshold) proceed to charging stations. The charging stations have a limited number of chargers, and each charger can only serve one robot at a time. After charging, the robot returns to the shelf storage area.
\item If the remaining battery level is above the threshold, the robot becomes idle and waits to be assigned to another order.
\end{enumerate}

Additionally, this paper makes the following assumptions: 1) The order arrival follows a Poisson distribution, and the arrival of varying multi-line orders is independent. The robot and order matching process follows a first-come, first-served (FCFS) principle. 2) The adopted dwell point policy, referred to as the point of service completion (POSC)~\cite{zou2018evaluating}, implies that robots are not required to return to a pre-established dwell point after completing an order. 3) A robot is assumed to execute one order at a time and will carry as many totes as possible during each trip to fulfill an order. 4) An order will not require a robot to retrieve a tote that has already been taken by another robot. This is based on the reality that a type of goods is typically distributed in numerous totes. 5) Robot velocity is constant; congestion and deadlock are not considered due to the unidirectional configuration of aisles and cross aisles. All robots are identical and have the same number of buffer positions.

\section{SHARED-TOKEN MULTI-CLASS SOQN} \label{sec: SOQN}

\subsection{MODEL DESCRIPTION}
\begin{figure}[htbp]
    \centering
    \includegraphics[width=0.45\textwidth]{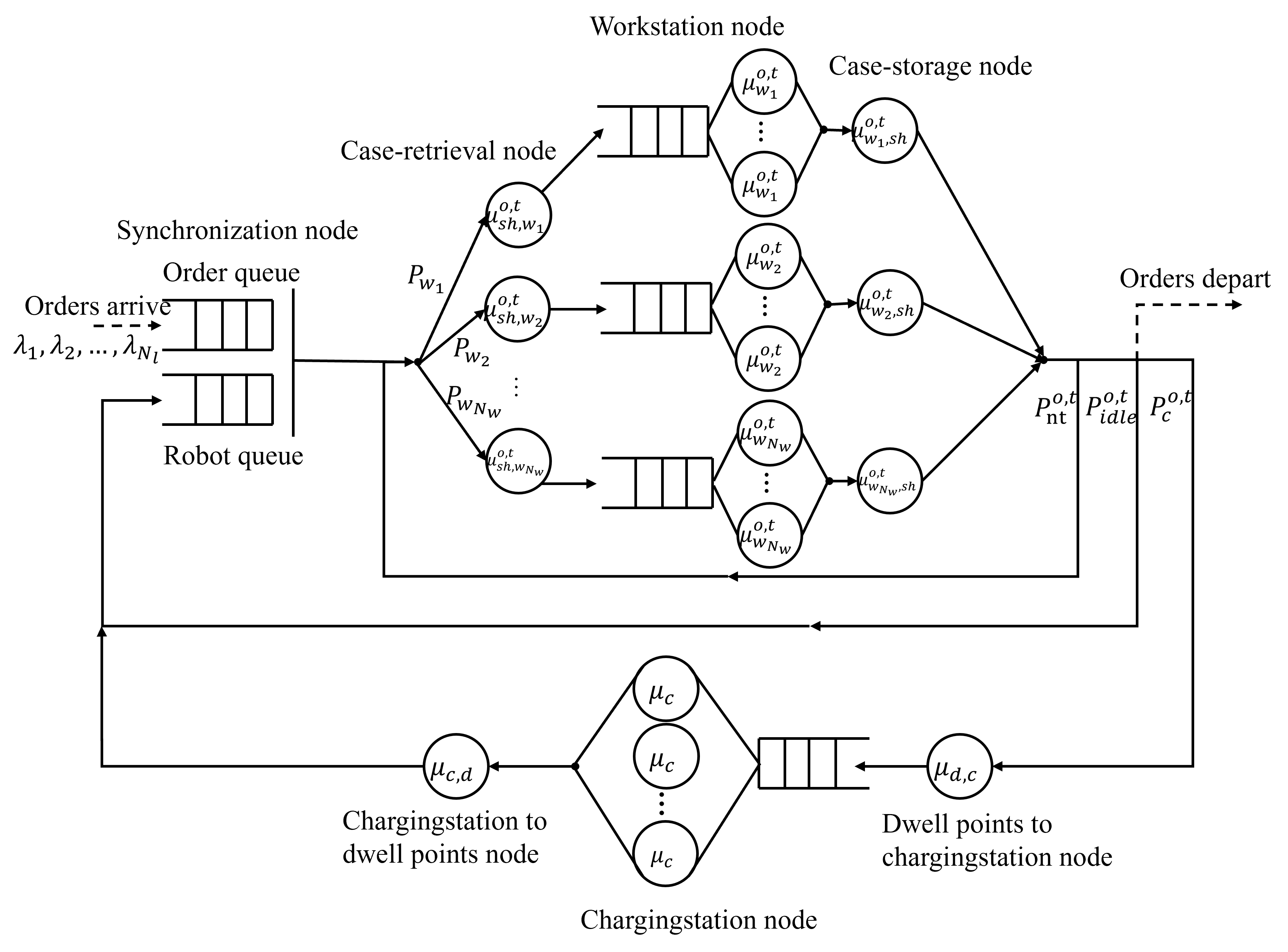}
    \vspace{-5pt}
    \caption{The shared-token, multi-class semi-open queueing network is constructed based on the MTSR system operation process.}
    \label{fig:SOQN}
    \vspace{-12pt}
\end{figure}

Fig.~\ref{fig:SOQN} illustrates the shared-token, multi-class SOQN model, which is constructed based on the operational processes of robots. The construction of the model and the explanation of notations are as follows:

\begin{table}[!ht]
\centering
\caption{SYMBOL DEFINITION}
\scalebox{1}{
\begin{tabularx}{0.47\textwidth}{>{\fontfamily{ptm}\selectfont}l >{\fontfamily{ptm}\selectfont}p{0.3\textwidth}}
\specialrule{1pt}{1pt}{1pt}
Symbol & Definition   \\ \specialrule{0.6pt}{1pt}{1pt}
$N_{sh}$ & The number of shelves in the warehouse \\
$sh_m, sh_n$ & The shelf with id $m$ and $n$\\
$v$ & The average velocity of the robot\\
$t_p$ & The average tote retrieval time from shelf\\
$th_c$ & The threshold for go charging \\
$dr$ & Battery depletion rate while moving\\
$\mathcal{O}=\{1,\dots,N_l\}$ & A set of indices of $N_l$ order classes\\
$\mathcal{W}=\{1,\dots,N_w\}$ & A set of indices of $N_w$ workstations   \\
$\mathcal{T}_o=\{1,\dots,NT_o\}$ & A set of indices of $NT_o$ trips needed to fulfill an $o$-class order\\
$N_r$ & The number of robots in the warehouse \\
$N_o$ & The number of lines within the class-$o$ order\\
$\lambda_{o}$ & The average arrival rate of the class-$o$ order\\
$\lambda$ & The overall average order arrival rate\\
$P_o$ & The probability that an arriving order belongs to the class-$o$ order \\
$NC_{o,t}$ & The number of totes required to be picked while executing class-$o$ order during trip $t$ \\
$C$ & The number of tote buffer positions on a robot\\
$NW_{i}$ & The number of workers in the workstation $i$\\
$\mu$ & The average service rate of nodes within the queueing network, which is the reciprocal of the average service time of a stage in the system's operation process\\
$T^{o,t}_{sh,w_i}, T^{o,t}_{w_i,sh}$ & Average time needed for a robot to retrieve totes required by class-$o$ order and transport to workstation $i$ during trip $t$, time required for the reverse storage process, $T^{o,t}_{sh,w_i}={\mu^{o,t}_{sh,w_i}}^{-1}, T^{o,t}_{w_i,sh}={\mu^{o,t}_{w_i,sh}}^{-1}$\\
$T^{o,t}_{w_i}$ & Average time needed for a worker in workstation $i$ to process all totes brought by the robot while executing $o$-class order in trip $t$, $T^{o,t}_{w_i}={\mu^{o,t}_{w_i}}^{-1}$ \\
$N_c$ & The number of chargers in the charging station \\
$T_c$ & Average robot charging time, $T_c={\mu_{c}}^{-1}$\\
$T_{c,d}, T_{c,d}$ & Average time required for a robot to travel from dwell points to charging station and from charging station back to its orginal position, $T_{c,d}={\mu_{c,d}}^{-1}$,$T_{d,c}={\mu_{d,c}}^{-1}$\\
$P_{w_i}$ & The probability to select workstation $i$ as target station\\
$P^{o,t}_{nt}, P^{o,t}_{c}, P^{o,t}_{idle}$ & The probability that the robot needs to continue proceeding to the next trip, going charging and becoming idle after trip $t$ while executing $o$-class order\\
$DB$ &  Average robot battery consumption per order\\
$ATT_{o,t}$ & The average travelling time for a robot to execute class-$o$ order in the $t^{th}$ trip\\
$TH$ & The maximum throughput of the warehouse \\
$THT, THT_o$ & The overall order throughput time and the value of $o$-class order\\
$\text{P}_{w_i}(n), \text{P}_{c}(n)$ & The probability of $n$ number of robots staying in
the workstation $i$ and charging station when there are $N_r$ robots in the warehouse\\
$NO_{sync}, NR_{sync}$ & The length of order queue and robot queue at the synchronization node\\
$\rho_{r},\rho_{c},\rho_{w}$ & The utilization rate of robots, chargers and workers\\
$WT_{w_i},WT_{c}$ & The average time the robot needs to wait to be served after arriving at the workstation $i$ and charging station\\
\specialrule{1pt}{1pt}{1pt}
\end{tabularx}}
\vspace{-20pt}
\end{table}

\boldpar{Synchronization Node}: In this network, there are $N_r$ robots and $N_l$ types of orders. Let $\mathcal{O}=\{1,\dots,N_l\}$ denote the set of indices for the $N_l$ order classes. For each $o \in \mathcal{O}$, the average order arrival rate is $\lambda_o$, and the number of lines within the order is defined as $N_o$. The probability of an order belonging to class-$o$ is $P_o = \frac{\lambda_o}{\sum_{j=1}^{N_l}\lambda_j}$. The matching process between robots and orders is modeled as a synchronization node, which features order and robot queues, with at least one queue always remaining empty~\cite{lamballais2017estimating}.

\boldpar{Case Retrieval Node}: 

Let $N_w$ denote the number of workstations in the logistics system, with $\mathcal{W}=\{1,\dots,N_w\}$ as the set of indices. If fulfilling a class-$o$ order requires $NT_o$ trips by the robot, the set of indices for all trips is denoted by $\mathcal{T}_o=\{1,\dots,NT_o\}$. Assuming the robot maximizes tote carriage per trip, $NT_o=\lceil\frac{N_o}{C}\rceil$. For each $o\in\mathcal{O}$ and $t\in\mathcal{T}_o$, the number of totes ($NC_{o,t}$) required to be picked during trip $t$ for a class-$o$ order is computed as follows:

\begin{equation}
    NC_{o,t} =
    \begin{cases}
    C\text{,} & \text{if } t<NT_o \\
    N_o-(t-1)C\text{,} & \text{if } t=NT_o
    \end{cases}
\end{equation}

The tote retrieval process involves visiting multiple shelves, picking the designated totes, and then transporting them to designated workstations. This process 
is modeled as an infinite service (IS) node~\cite{bolch2006queueing} with an average service rate of $\mu_{sh,w_i}^{o,t}$. 
This is because once the robot is matched with an order, it can commence its travel immediately without any waiting time. The average travel time for each order $o\in \mathcal{O}$, trip $t\in \mathcal{T}_o$, and workstation $i\in \mathcal{W}$ is represented by $\frac{1}{\mu_{sh,w_i}^{o,t}}$.

The travel time during trip $t$ is heavily influenced by the tote retrieval sequence policy. For the random retrieval sequence policy, the travel distances for retrieving a tote within a trip is i.i.d. The average travel time, denoted as $T_{sh,w_i}^{o,t}$, is calculated by:
\begin{align}
    \begin{split}
        T_{sh,w_i}^{o,t} = 
        &\frac{\sum_{m=1}^{N_{sh}}\text{D}(sh_m,w_i)}{N_{sh}v} \\
        & +NC_{o,t}\cdot\left(\frac{\sum_{m=1}^{N_{sh}}\sum_{n=1}^{N_{sh}}\text{D}(sh_m,sh_n)}{N_{sh}^2v}+t_p\right)\text{,}
    \end{split}
\end{align}

where $\text{D}(\cdot)$ refers to the shortest traveling distance between two points, adhering to the unidirectional path network. Given that the layout considered in this work is similar to that studied in the RMFS research~\cite{lamballais2017estimating}, the method for calculating the distance between two points can be similarly applied in our case. Assume the warehouse contains $N_{sh}$ shelves, with $sh_m$ and $sh_n$ denoting the $m$-th and $n$-th shelf, respectively. The time a robot requires to pick a tote from a shelf is denoted as $t_p$.

Under the CR policy, the nearest neighbor heuristic~\cite{qin2024performance} makes the travel distance required to retrieve a tote interdependent with each other. For example, the first retrieved tote within a trip is not uniformly distributed among different shelves due to the nearest neighbor rule. The position of the first tote becomes the starting point for retrieving the second tote, thereby causing the travel distance to retrieve the second tote influenced by the first tote. The total possible combinations of tote locations for each class-$o$ order is $N_{sh}^{N_{o}}$. Therefore, calculating the average travel distances for class-$o$ orders in different trips analytically requires the analysis of every possible case, resulting in a computational complexity exceeding $O(N_{sh}^{NC_{o,t}})$. Hence, it is not scalable to conduct an exact analytical analysis. Instead of using the numerical method discussed in~\cite{qin2024performance}, we use the Monte Carlo method~\cite{metropolis1949monte} to estimate the average travel time through random sampling from discrete-event simulation. This method is more accurate and applicable to different warehouse layouts as long as the number of samples is sufficient. Additionally, since congestion is not considered due to the unidirectional path setting, the number of robots will not influence the average travel time. Therefore, as long as the warehouse layout does not change, the average travel time remains fixed. During sampling, the simulation can stop once the 95\% confidence interval of the average travel time is within 1\% of its mean.

\boldpar{Workstation Node}: For each $i \in \mathcal{W}$, the tote handling process at workstation $i$ is modeled as a service node with $NW_i$ servers. If the time required for a worker to handle a tote is a random variable denoted as $X_c$, then the time to handle all totes brought by a robot while executing an $o$-class order over $t$ trips is equal to $\sum_{j=1}^{NC_{o,t}} X_c^j$. If every tote handling time is i.i.d. and follows a uniform distribution ($U[a, b]$), then the mean and the squared coefficient of variation for all tote handling times can be calculated as follows:
\begin{equation}
    \frac{1}{\mu_{w_i}^{o,t}}=\mathbb{E}[\sum^{NC_{o,t}}_{j=1}X_c^j]=\sum^{NC_{o,t}}_{j=1}\mathbb{E}[X_c^j]=\frac{a+b}{2}\cdot NC_{o,t}\text{, } 
    \label{eqn:uniform distribution}
\end{equation}

\begin{align}
    {cv_{w_i}^{o,t}}^2 &= \frac{\mathbb{E}\left[\left(\sum_{j=1}^{NC_{o,t}} X_c^j\right)^2\right] - \mathbb{E}\left[\sum_{j=1}^{NC_{o,t}} X_c^j\right]^2}{\mathbb{E}\left[\sum_{j=1}^{NC_{o,t}} X_c^j\right]^2} \\
    &= \frac{1}{NC_{o,t}} \cdot \frac{\mathbb{E}[X_c^2] - \mathbb{E}[X_c]^2}{\mathbb{E}[X_c]^2} \nonumber\\
    &= \frac{1}{NC_{o,t}} \cdot \frac{(b-a)^2}{3(a+b)^2} \nonumber
\end{align}

\boldpar{Case Storage Node}: Storing totes back to their original positions after service completion at a workstation is also modeled as an IS node with the service rate set at $\mu_{w_i,sh}^{o,t}$. The average travel time ($\frac{1}{\mu_{w_i,sh}^{o,t}}$) could be calculated in a manner similar to the tote retrieval stage.

\boldpar{Routing Probability}:
For each $i\in\mathcal{W}$, the probability $P_{w_i}$ of a robot choosing workstation $i$ is proportional to its worker count, expressed as $P_{w_i}=\frac{NW_{i}}{\sum_{i=1}^{N_w}NW_{i}}$, with $NW_{i}$ denoting the number of workers at station $i$.

After completing the tote storage process, the robot has a probability of $P_{nt}^{o,t}$ to proceed to the next trip following trip $t$, beginning with the tote retrieval process. The probability $P_{nt}^{o,t}$ is calculated as follows:
\begin{equation}
    P^{o,t}_{nt} = 
    \begin{cases}
        1\text{,} & \text{if } t<NT_{o} \\
        0\text{,}  & \text{if } t=NT_{o}
    \end{cases}
\end{equation}
With a probability of $P_c^{o,t}$, the robot needs to go charging after finishing trip $t$ of the $o$-class order. When a robot finishes an order, the probability of going charging is the reciprocal of the average number of orders that a fully charged battery can support before going to charge, which could be estimated as follows:
\begin{equation}
    P^{o,t}_{c} \approx 
    \begin{cases}
        \frac{DB}{100-th_c}\text{,} & \text{if } t=NT_{o} \\
        0\text{,} & \text{otherwise,}
    \end{cases}
\end{equation}
where $DB$ refers to the average battery consumption for a robot to fulfill an order. Assuming that the battery consumption is linearly related to the travel time, it can be calculated as follows:
\begin{equation}
    DB=\Xsum_{o=1}^{N_l} P_o \Xsum_{t=1}^{NT_o} dr\cdot ATT_{o,t} \text{,}
\end{equation}
where $dr$ denotes the percentage of battery depletion rate per minute while moving; $ATT_{o,t}$ refers to the average travel time for a robot to execute a class-$o$ order on the $t^{th}$ trip.  
\begin{equation}
    ATT_{o,t} = \Xsum_{i=1}^{N_w}P^o_{w_i}(\frac{1}{\mu_{sh,w_i}^{o,t}}+ \frac{1}{\mu_{w_i,sh}^{o,t}})
\end{equation}
For $o \in \mathcal{O}, t \in \mathcal{T}_o$, the probability that a robot becomes idle and moves to the robot queue at the synchronization node after completing trip $t$ of order $o$ is:
\begin{equation}
    P^{o,t}_{idle}=1-P^{o,t}_{c}-P^{o,t}_{nt}
\end{equation}

\boldpar{Charging related nodes}: Robots traveling from their dwelling points to the charging station and from the charging station back to their dwelling points are also modeled as IS nodes, with average traveling time of $\frac{1}{\mu_{d,c}}$ and $\frac{1}{\mu_{c,d}}$, respectively. Given that each shelf has an equal chance of being designated as the dwelling point under a random-tote retrieval policy, the traveling time is calculated as follows:
\begin{equation}
    \frac{1}{\mu_{d,c}}\approx\frac{1}{\mu_{c,d}}=\frac{\Xsum_{m=1}^{N_{sh}}D(sh_m,c)}{N_{sh}v}
\end{equation}
For the CR policy, determining the average travel time also requires simulation, similar to the retrieval time calculation. 

Suppose there is only one charging station in the warehouse, equipped with $N_c$ servers. The robot charging process can be modeled as a service node with $N_c$ servers. Assuming the charging time ($t_c$) follows a uniform distribution $U(c, d)$, the average charging time ($\frac{1}{u_c}$) and its coefficient of variation ($cv_c$) at the charging station can be calculated according to Eqn.7 in~\cite{zou2018evaluating}.

\subsection{SOLUTION APPROCH}
The SOQN model for the MTSR system is solvable through single-chain, multiclass AMVA~\cite{buitenhek2000amva}. This requires converting SOQN routing probabilities into normalized visit ratios ($V^{o,t}_{node}$) that indicate the relative frequency of robot visits to network nodes while executing $o$-class order in trip $t$~\cite{bolch2006queueing}. To normalize these ratios, we could set the robot visit ratio of the synchronization node to $1$:
\begin{equation}
    V_{sync}=\Xsum_{o=1}^{N_l}\Xsum_{t=1}^{NT_o}V^{o,t}_{sync}=\Xsum_{o=1}^{N_l}P_o=1 \text{,}
\end{equation}
The normalized visit ratio for a robot serving an $o$-class order on its $t^{th}$ trip to the tote retrieval, workstation, and tote storage nodes associated with workstation $i$, for each $o \in \mathcal{R}, t \in \mathcal{T}_o, i \in \mathcal{W}$, is calculated as follows:
\begin{equation}
    V_{sh,w_i}^{o,t}=V_{w_i}^{o,t}=V_{w_i,sh}^{o,t}=P_{w_i}P_o
\end{equation}
While the robot goes for charging, it does not serve any orders. Therefore, the normalized visit ratio to the charging-related nodes can be calculated as follows:
\begin{equation}
    V_{c}=V_{c,d}=V_{d,c}=\Xsum_{o=1}^{N_l}P_oP_c^{o,NT_o}
\end{equation}

According to~\cite{buitenhek2000amva}, the solution to the proposed shared-token, multi-class SOQN model follows three steps:

\boldpar{Step 1}: A Closed QN (CQN) is formed by removing the synchronization station from the SOQN. This network is analyzed using the single-chain, multi-class AMVA~\cite{buitenhek2000amva}. The maximum system throughput ($TH$) with $N_r$ robots operating in the network can be calculated using AMVA.

\boldpar{Step 2}: The overall average order arrival rate is \(\lambda = \sum_{o=1}^{N_l} \lambda_o\). If the maximum system throughput is less than the overall order arrival rate, orders will continuously accumulate, leading to an unstable system.

If the system is stable, a second CQN is formed by replacing the synchronization node in the SOQN with a load-dependent service node, where the service rate depends on the number of idle robots in the system. The service rate is $\lambda=\sum_{o=1}^{N_l}\lambda_o$ when there are more than one idle robot. When there is only one idle robot, it becomes $\frac{\lambda \cdot TH}{TH-\lambda}$.   The AMVA algorithm is employed once again to analyze this second CQN, taking into account the influence of arriving orders. The analysis outputs include the average time a robot must wait to be served after arriving at the charging station ($WT_{c}$) and at different workstations ($WT_{w_i}$), the expected number of robots queuing at the synchronization node ($NR_{sync}$), and the probability of $n$ robots being present at different workstations ($P_{w_i}(n)$) or at the charging station ($P_{c}(n)$), given $N_r$ robots in the warehouse.

\boldpar{Step 3}: 
This step involves isolating the synchronization node to calculate the mean order queue length ($NO_{sync}$), which refers to the average number of arriving orders waiting to be assigned to idle robots.

The order throughput time and the utilization rate of resources can be derived from \textbf{Step 2} and \textbf{Step 3}. Robot utilization ($\rho_r$) is computed as the percentage of busy robots not available for assignment: $\rho_r = (1 - \frac{NR_{\text{sync}}}{N_r}) \times 100\%$. The utilization rate of workers can be calculated as: $\rho_w = \sum_{i=1}^{N_w} P_{w_i} \left(1 - \sum_{n=0}^{NW_i-1} \frac{NW_i-n}{NW_i} P_{w_i}(n)\right) \times 100\%$. Given that there are $N_c$ chargers in the charging station, the utilization rate of chargers can be calculated as $\rho_c = (1 - \sum_{n=0}^{N_c-1} \frac{N_c-n}{N_c} P_c(n)) \times 100\%$. The order throughput time, $THT_o$, calculates the duration from when a class-$o$ order arrives to when it is completed, taking into account the external order waiting time, the average robot travel time for tote retrieval and storage processes, and the waiting and service time at different workstations. It can be calculated based on Eqn.~\ref{eqn:THT_o}. Overall order throughput is calculated accordingly in Eqn.~\ref{eqn:THT}.

\begin{equation}
    \begin{split}
    THT_o=
    & \Xsum_{t=1}^{T_o}\Xsum_{i=1}^{N_w}P^o_{w_i}(\frac{1}{\mu_{sh,w_i}^{o,t}}+\frac{1}{\mu_{w_i}^{o,t}}+\frac{1}{\mu_{w_i,sh}^{o,t}}+WT_{w_i})\\
    & +\frac{NO_{sync}}{\lambda}
    \label{eqn:THT_o}
    \end{split}
\end{equation}
\vspace{-10pt}
\begin{equation}
    THT =\Xsum_{o=1}^{N_l}P_o\cdot THT_o
    \label{eqn:THT}
\end{equation}

\section{NUMERICAL EXPERIMENTS} \label{sec: numerical experiments}
This section validates the proposed QN model through the discrete event simulation, examining the number of robots, tote buffer positions, and the effects of CR and random tote retrieval policies on the system's steady-state performance.

\subsection{Experimental setup}

\begin{figure}[!ht]
    \centering
    \includegraphics[width=0.48\textwidth]{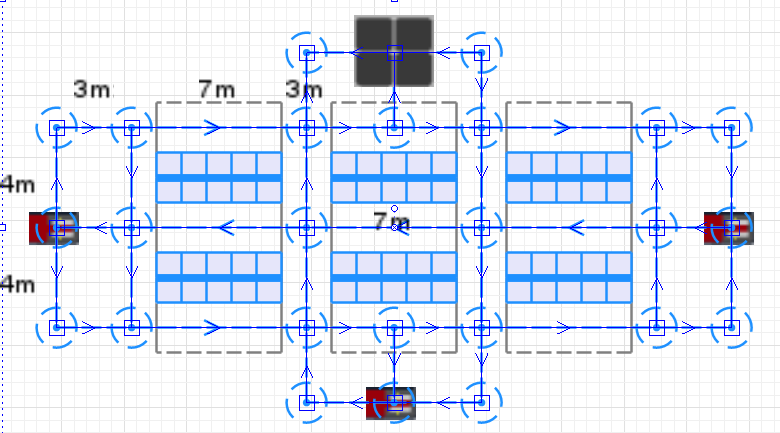}
    \vspace{-5pt}
    \caption{A small warehouse layout with unidirectional paths has three workstations on the west, south, and east sides of the warehouse, with the charging station located at the top.}
    \label{fig:experiment layout}
    \vspace{-10pt}
\end{figure}

In Sec.~\ref{subsec:results and analysis}, we conduct three experiments to analyze the MTSR system in a small warehouse layout, as depicted in Fig.~\ref{fig:experiment layout}. In the workstation, the time required for a worker to handle a tote follows a uniform distribution of $U(5,8)$ seconds. We set the maximum number of lines within an order to five, and the probabilities that an arriving order will require one to five lines are set to $[0.1,~0.2,~0.3,~0.2,~0.2]$, respectively. The average velocity of the robot is $0.5$ m/s, the average tote retrieval time from the shelf is five seconds, the charging threshold is 20\%, and the battery depletion rate is set to 0.5\% per minute while moving, the distribution of robot charging time is $U(25,35)$ minutes.


The discrete event simulation is programmed based on the agent behavior logic illustrated in Sec.~\ref{sec: System description} in AnyLogic software (version 8.8.6)~\cite{AnyLogicSimulation}. For the experiments shown below, each scenario is simulated using 20 replications with different random seeds, where each replication lasts 1,000 simulation hours. This results in a 95\% confidence interval where the half-width is within 1\% of the average performance metric. These metrics include order throughput time ($THT$) and utilization of resources ($\rho_r, \rho_w, \rho_c$). The accuracy of the QN models is assessed by the absolute relative error rate, denoted as $\delta = \frac{|A - S|}{A} \times 100\%$, where $A$ and $S$ represent the QN model and simulation result, respectively.


\subsection{Results and Analysis}\label{subsec:results and analysis}
\boldpar{Varying number of robots:}
We first investigate how varying the number of robots influences the system performance, with a fixed number of buffer positions on robots ($C = 4$), and the overall average order arrival rate of $2$ orders/min. Each workstation has $1$ worker, and there are $4$ chargers at the charging station.

\begin{figure*}[t]   
    \subfloat[\label{fig:performance vs robot}]{%
      \begin{minipage}[t]{0.31\linewidth}
        \centering 
        \includegraphics[width=2.3in]{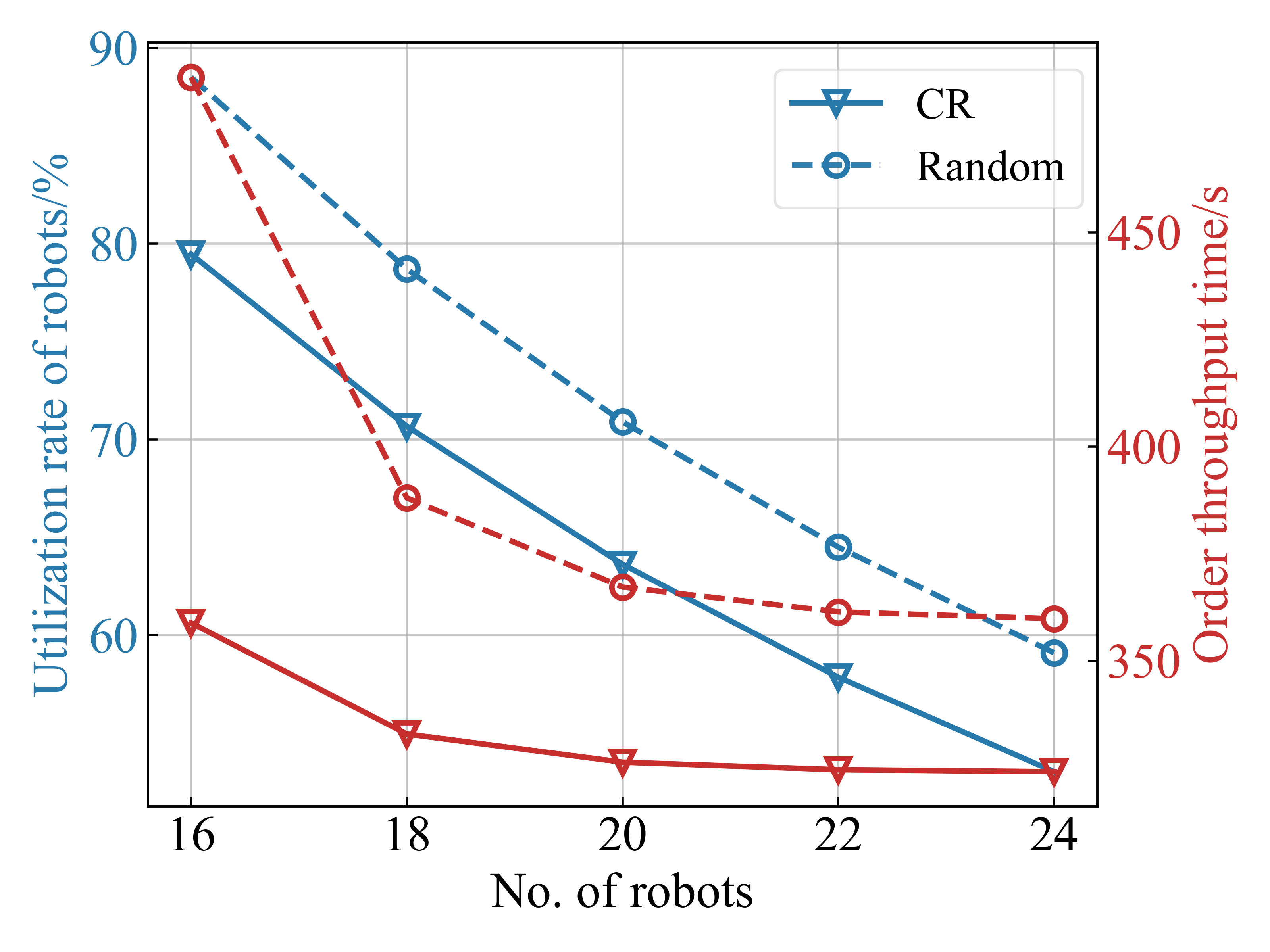}   
      \end{minipage}%
      }
      \hfill
    \subfloat[\label{fig:performance vs capacity}]{%
      \begin{minipage}[t]{0.31\linewidth}   
        \centering   
        \includegraphics[width=2.3in]{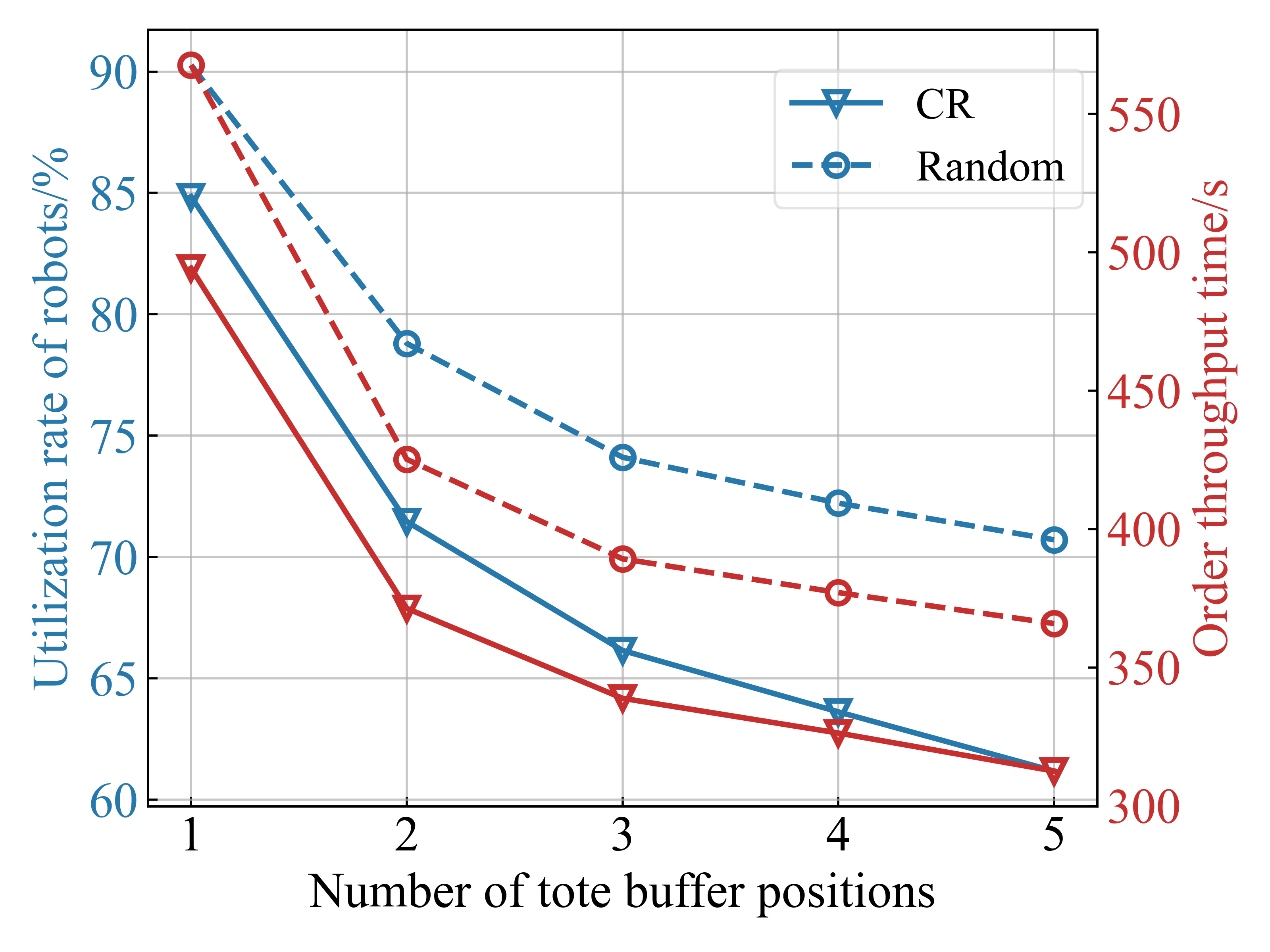}   
      \end{minipage} 
      }
       \hfill
    \subfloat[\label{fig:performance vs order}]{%
      \begin{minipage}[t]{0.31\linewidth}   
        \centering   
        \includegraphics[width=2.3in]{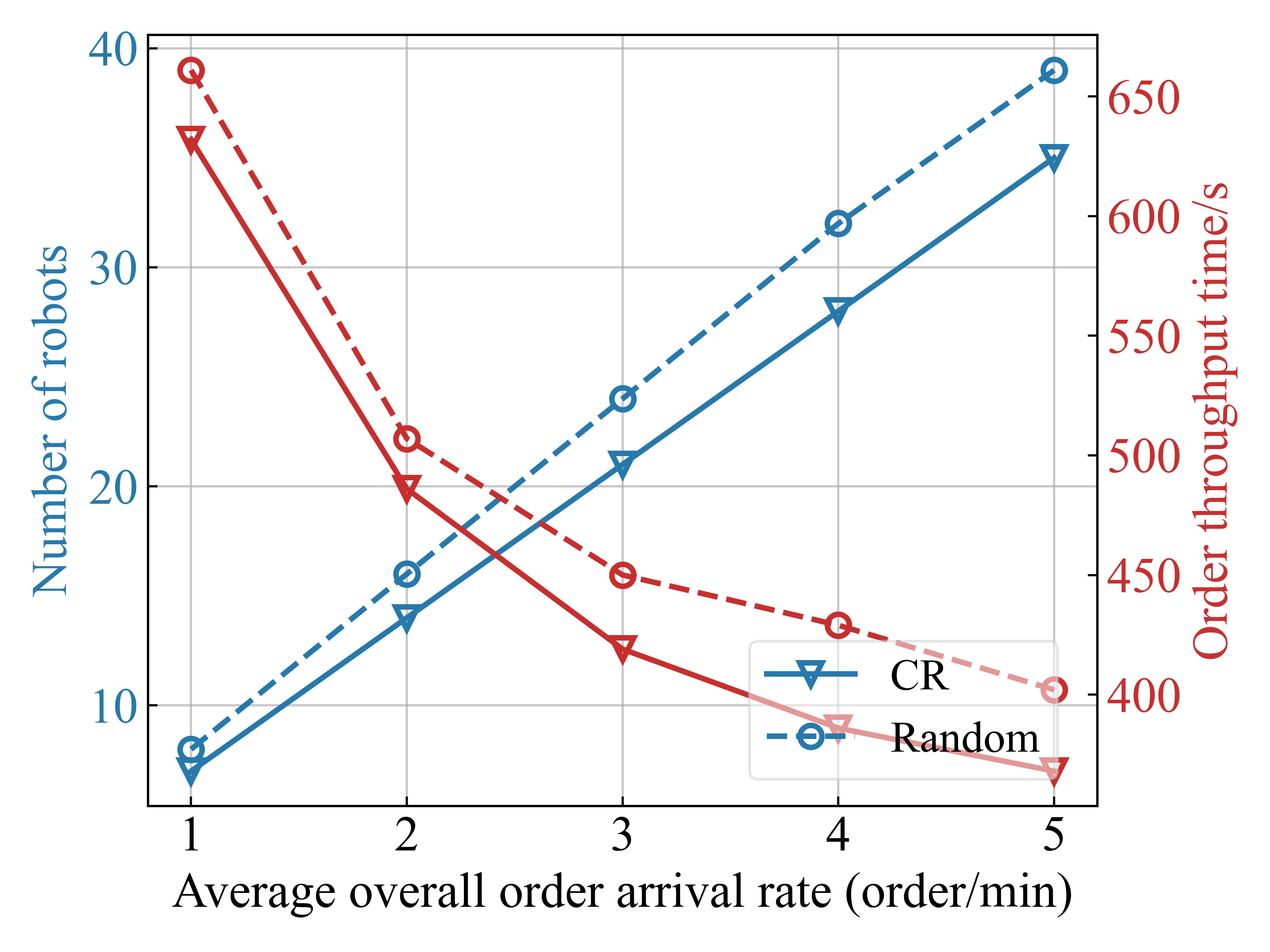}   
      \end{minipage} 
    }
    \vspace{-0.1cm}
    \caption{Under both CR and random tote retrieval policies, we compare:
(a) Robot utilization and order throughput time with varying numbers of robots in the system.
(b) Robot utilization and order throughput time with varying buffer positions.
(c) Minimum number of robots needed to maintain system stability and corresponding order throughput time with varying average order arrival rate.} 
    \label{fig:comparsions}
    \vspace{-0.1cm}
\end{figure*}

Fig.~\ref{fig:performance vs robot} displays the steady-state system performance as the number of robots varies under both random and CR tote retrieval policies. As the number of robots increases, the utilization rate of the robots decreases. This is because the maximum system throughput increases while the order arrival rate remains fixed, resulting in a growing number of idle robots awaiting order allocation. The trend of decreasing order throughput time is highly nonlinear and gradually levels off as the number of robots increases. Additionally, when the number of robots in the system exceeds $20$, the difference in the order throughput time between the two policies stabilizes. This is because, with an increasing number of robots, the order waiting time for an idle robot gradually converges to $0$ for both policies. The primary difference then lies in the travel time during the tote retrieval and storage process, which remains consistent regardless of the number of robots. The comparison results with the simulation are shown in Table~\ref{table: 2}.

\boldpar{Varying tote buffer positions:}
We then examine how the robot's tote buffer positions affect system performance with a fixed number of 20 robots.
Other parameters, like the average order arrival rate, workers and charger counts, remain consistent with the previous experiment.

In Fig.~\ref{fig:performance vs capacity}, increasing buffer positions reduces both the order throughput time and robot utilization, due to fewer round trips between the shelf storage area and the workstation. However, the performance gains achieved by increasing buffer positions diminish as buffer positions increase, since the number of trips a robot must make to fulfill an order does not decrease linearly with increased buffer positions. For instance, when buffer positions increase from $1$ to $2$, the trips needed to fulfill an order with $4$ lines reduce from $4$ to $2$. However, increasing the buffer positions from $2$ to $3$ does not change the trip count. After comparing with the simulation results, the average relative error ratios for the order throughput time, utilization of robots, chargers, and workers are found to be $1.4\%$, $0.8\%$, $1.1\%$, and $0\%$, respectively (see Table~\ref{table: 1} for details).

\boldpar{Varying average order arrival rate:}
We further explore the minimum number of robots required to maintain system stability, with the average resource (robots, chargers, workers) utilization rates below $90\%$ across varying order arrival rates. Setting a $90\%$ threshold accounts for contingencies in system operation, such as robot paralysis. Workers are evenly distributed among workstations, with at most one-worker difference. Because the computation of the QN-based performance estimation is inexpensive, we use brute force optimization to maximize robot utilization. The number of buffer positions on a robot is $4$.

As shown in Fig.\ref{fig:performance vs order}, the required minimum number of robots increases with the average order arrival rate. The CR policy requires $12.5\%$ fewer robots to guarantee the system stable than the random policy under the same average order arrival rate. As for order throughput time, the CR policy performs better than the random policy, due to lower travel times and shorter order queues waiting for idle robots. As the order arrival rate increases, the order throughput time appears to decrease, while the robot utilization rate remains around 90\%. The simulation validation results are shown in Table~\ref{table: 3}.


\begin{table*}[!ht]
\centering
\caption{Validation results with varying number of robots}
\scalebox{1}{
\begin{tabular}{cccccccccccccc}
\hline
                    &        & $\rho_{r}$(\%) &      &       & $\rho_{w}$(\%) &      &       & $\rho_{c}$(\%) &      &       & THT(s) &       &       \\ \hline
$N_r$                & Policy & A          & S    & $\delta$(\%) & A          & S    & $\delta$(\%) & A          & S    & $\delta$(\%) & A      & S     & $\delta$(\%) \\ \hline
\multirow{2}{*}{16} & Random & 88.5       & 89.2 & 0.8   & 23.1       & 23.1 & 0.0   & 51.7       & 51.8 & 0.2   & 486.2  & 461.8 & 5.0   \\
                    & CR & 79.5       & 79.1 & 0.5   & 23.1       & 23.1 & 0.0   & 45.4       & 46.0 & 1.3   & 358.8  & 350.4 & 2.3   \\
\multirow{2}{*}{18} & Random & 78.7       & 79.2 & 0.6   & 23.1       & 23.1 & 0.0   & 51.7       & 51.8 & 0.2   & 388.9  & 388.2 & 0.2   \\
                    & CR & 70.7       & 70.5 & 0.3   & 23.1       & 23.1 & 0.0   & 45.4       & 46.8 & 3.1   & 332.9  & 330.0 & 0.9   \\
\multirow{2}{*}{20} & Random & 70.9       & 71.6 & 1.0   & 23.1       & 23.1 & 0.0   & 51.7       & 51.8 & 0.2   & 367.2  & 372.5 & 1.4   \\
                    & CR & 63.6       & 63.3 & 0.5   & 23.1       & 23.1 & 0.0   & 45.4       & 46.2 & 1.8   & 326.3  & 325.4 & 0.3   \\
\multirow{2}{*}{22} & Random & 64.5       & 65.3 & 1.2   & 23.1       & 23.1 & 0.0   & 51.7       & 51.8 & 0.2   & 361.4  & 367.6 & 1.7   \\
                    & CR & 57.8       & 58.5 & 1.2   & 23.1       & 23.1 & 0.0   & 45.4       & 46.5 & 2.4   & 324.6  & 323.9 & 0.2   \\
\multirow{2}{*}{24} & Random & 59.1       & 59.7 & 1.0   & 23.1       & 23.1 & 0.0   & 51.7       & 51.8 & 0.2   & 359.8  & 364.8 & 1.4   \\
                    & CR & 53.3       & 53.7 & 0.8   & 23.1       & 23.1 & 0.0   & 45.4       & 46.2 & 1.8   & 321.1  & 324.0 & 0.9   \\ 
\textbf{Average} &&&& \textbf{0.8} &&& \textbf{0} &&& \textbf{1.1} &&& \textbf{1.4} \\
                   \hline
\end{tabular}}
\label{table: 2}
\end{table*}

\begin{table*}[!ht]
\centering
\caption{Validation results with varying tote buffer positions}
\scalebox{1}{
\begin{tabular}{cccccccccccccc}
\hline
                   &        & $\rho_{r}$(\%) &      &       & $\rho_{w}$(\%) &      &       & $\rho_{c}$(\%) &      &       & THT(s) &       &       \\ \hline
C                  & Policy & A          & S    & $\delta$(\%) & A          & S    & $\delta$(\%) & A          & S    & $\delta$(\%) & A      & S     & $\delta$(\%) \\ \hline
\multirow{2}{*}{1} & Random & 90.3       & 89.1 & 1.3   & 23.1       & 23.1 & 0.0   & 65.2       & 65.1 & 0.2   & 567.7  & 535.1 & 5.7   \\
                   & CR & 84.8       & 84.4 & 0.5   & 23.1       & 23.1 & 0.0   & 62.5       & 62.6 & 0.2   & 494.3  & 470.6 & 4.8   \\
\multirow{2}{*}{2} & Random & 78.8       & 78.1 & 0.9   & 23.1       & 23.1 & 0.0   & 57.7       & 56.6 & 1.9   & 425.2  & 413.6 & 2.7   \\
                   & CR & 71.5       & 71.4 & 0.1   & 23.1       & 23.1 & 0.0   & 51.7       & 52.5 & 1.5   & 371.4  & 370.6 & 0.2   \\
\multirow{2}{*}{3} & Random & 74.1       & 73.1 & 1.3   & 23.1       & 23.1 & 0.0   & 53.6       & 53.1 & 0.9   & 389.2  & 381.3 & 2.0   \\
                   & CR & 66.2       & 65.8 & 0.6   & 23.1       & 23.1 & 0.0   & 48.4       & 47.5 & 1.9   & 338.9  & 338.7 & 0.1   \\
\multirow{2}{*}{4} & Random & 72.2       & 71.7 & 0.7   & 23.1       & 23.1 & 0.0   & 51.7       & 51.9 & 0.4   & 377.1  & 370.5 & 1.8   \\
                   & CR & 63.6       & 63.6 & 0.0   & 23.1       & 23.1 & 0.0   & 45.4       & 46.5 & 2.4   & 326.3  & 325.5 & 0.2   \\
\multirow{2}{*}{5} & Random & 70.7       & 69.9 & 1.1   & 23.1       & 23.1 & 0.0   & 51.7       & 50.7 & 1.9   & 365.9  & 360.1 & 1.6   \\
                   & CR & 61.2       & 61.1 & 0.2   & 23.1       & 23.1 & 0.0   & 44.1       & 44.1 & 0.0   & 315.8  & 312.6 & 1.0   \\ 
\textbf{Average} &&&& \textbf{0.7} &&& \textbf{0} &&& \textbf{1.2} &&& \textbf{2.0} \\
                   \hline

\end{tabular}}
\label{table: 1}
\end{table*}

\begin{table*}[!ht]
\centering
\caption{Validation results with varying average order arrival rate}
\scalebox{0.98}{
\begin{tabular}{ccccccccccccccccc}
\hline
                   &        &         &         &         & THT(s) &       &       & $\rho_{r}$(\%) &      &       & $\rho_{c}$(\%) &      &       & $\rho_{w}$(\%) &       &       \\ \hline
$\lambda$          & Policy & $N_{r}$ & $N_{c}$ & $N_{w}$ & A      & S     & $\delta$(\%) & A          & S    & $\delta$(\%) & A          & S    & $\delta$(\%) & A          & S     & $\delta$(\%) \\ \hline
\multirow{2}{*}{1} & Random & 8.0     & 2.0     & 3.0     & 661.5  & 584.8 & 11.6  & 89.2       & 89.3 & 0.1   & 52.3       & 52.9 & 0.2   & 11.5       & 11.6 & 0.8   \\
                   & CR & 7.0     & 3.0     & 4.0     & 632.1  & 586.1 & 7.3   & 89.5       & 90.4 & 1.0   & 31.3       & 30.7 & 1.3   & 8.8        & 8.7  & 1.3   \\
\multirow{2}{*}{2} & Random & 16.0    & 3.0     & 3.0     & 507.7  & 470.2 & 7.4   & 89.5       & 89.6 & 0.1   & 69.3       & 69.2 & 0.2   & 23.2       & 23.1 & 0.4   \\
                   & CR & 14.0    & 4.0     & 5.0     & 486.2  & 433.8 & 10.8  & 89.6       & 89.7 & 0.1   & 46.0       & 45.9 & 3.1   & 13.8       & 13.8 & 0.0   \\
\multirow{2}{*}{3} & Random & 24.0    & 5.0     & 3.0     & 450.1  & 433.3 & 3.7   & 89.1       & 89.8 & 0.7   & 63.3       & 62.3 & 0.2   & 34.5       & 34.6 & 0.2   \\
                   & CR & 21.0    & 7.0     & 5.0     & 419.8  & 391.8 & 6.7   & 89.8       & 90.4 & 0.6   & 39.7       & 39.4 & 1.8   & 20.6       & 20.8 & 0.9   \\
\multirow{2}{*}{4} & Random & 32.0    & 5.0     & 4.0     & 429.4  & 422.5 & 1.6   & 89.7       & 90.3 & 0.6   & 83.3       & 83.6 & 0.2   & 34.8       & 34.8 & 0.0   \\
                   & CR & 28.0    & 7.0     & 6.0     & 386.8  & 363.8 & 5.9   & 89.3       & 89.7 & 0.4   & 52.6       & 52.3 & 2.4   & 23.3       & 23.1 & 0.9   \\
\multirow{2}{*}{5} & Random & 39.0    & 11.0    & 6.0     & 402.8  & 400.3 & 0.6   & 89.9       & 90.7 & 0.8   & 48.0       & 47.2 & 0.2   & 28.7       & 28.9 & 0.7   \\
                   & CR & 35.0    & 8.0     & 7.0     & 368.1  & 360.0 & 2.2   & 89.8       & 90.4 & 0.7   & 57.7       & 57.5 & 1.8   & 24.1       & 24.5 & 1.6   \\ 
\textbf{Average} & & &&&&& \textbf{5.8} &&& \textbf{0.5} &&& \textbf{1.2} &&& \textbf{0.7} \\
                   \hline

\end{tabular}}
\label{table: 3}
\end{table*}

\section{ CONCLUSIONS AND FUTURE WORK} \label{sec: conclusion}
In this paper, we developed a shared-token, multi-class SOQN to assess the performance of an MTSR system, accounting for multi-line orders with general distribution forms. Orders with different numbers of lines are categorized into distinct classes due to their varying time distributions spent at different stages of the operation process. The SOQN model is solved using the AMVA, and validated through simulations with most key performance metrics demonstrating an average accuracy of over 98\%. In our experimental setting, results indicate a 12.5\% reduction in the minimum number of robots needed to satisfy a specific order arrival rate using the CR policy compared with the random policy. The relationship between the number of robots, tote buffer positions, and order throughput time is nonlinear. The proposed model could be highly useful for expediting steady-state performance prediction and optimizing resource specifications for tailored warehouses in the pre-deployment stage. This enables warehouse owners to identify resource bottlenecks and manage budgets more effectively. An example application, "QN-based Warehouse Consultation Tool," is demonstrated in the demo video\footnote{\url{https://drive.google.com/drive/folders/1uAo96fgU91stykFx0rgzeS5XelB_TJk7?usp=sharing}}.

Currently, the effect of aisle blocking by robots while retrieving totes from high shelves is often overlooked. As the number of robots operating within the warehouse increases, leading to more congested conditions, the impact of blocking on overall system performance becomes more significant. Therefore, it merits detailed investigation in future work. Furthermore, certain shelves within the MTSR system are multi-deep, requiring a robot to remove the outer tote before accessing an inner one. This process warrants inclusion in future models. Additionally, some warehouses designate specific regions for robots to dwell in when they become idle, rather than having them dwell at their last operational location. Investigating the efficacy of this dwell point policy on warehouse performance is also of considerable interest.





\bibliographystyle{IEEEtran}
\bibliography{reference}

\begin{thebibliography}{10}
\providecommand{\url}[1]{#1}
\csname url@samestyle\endcsname
\providecommand{\newblock}{\relax}
\providecommand{\bibinfo}[2]{#2}
\providecommand{\BIBentrySTDinterwordspacing}{\spaceskip=0pt\relax}
\providecommand{\BIBentryALTinterwordstretchfactor}{4}
\providecommand{\BIBentryALTinterwordspacing}{\spaceskip=\fontdimen2\font plus
\BIBentryALTinterwordstretchfactor\fontdimen3\font minus \fontdimen4\font\relax}
\providecommand{\BIBforeignlanguage}[2]{{%
\expandafter\ifx\csname l@#1\endcsname\relax
\typeout{** WARNING: IEEEtran.bst: No hyphenation pattern has been}%
\typeout{** loaded for the language `#1'. Using the pattern for}%
\typeout{** the default language instead.}%
\else
\language=\csname l@#1\endcsname
\fi
#2}}
\providecommand{\BIBdecl}{\relax}
\BIBdecl

\bibitem{teck2023efficient}
S.~Teck, P.~Vansteenwegen, and R.~Dewil, ``An efficient multi-agent approach to order picking and robot scheduling in a robotic mobile fulfillment system,'' \emph{Simul. Model. Pract. Theory.}, p. 102789, 2023.

\bibitem{justkowiak2023stronger}
J.-E. Justkowiak and E.~Pesch, ``Stronger mixed-integer programming-formulations for order-and rack-sequencing in robotic mobile fulfillment systems,'' \emph{Eur. J. Oper. Res}, 2023.

\bibitem{lu2023automated}
J.~Lu \emph{et~al.}, ``An automated guided vehicle conflict-free scheduling approach considering assignment rules in a robotic mobile fulfillment system,'' \emph{CAIE}, vol. 176, p. 108932, 2023.

\bibitem{shan2024distributed}
X.~Shan \emph{et~al.}, ``A distributed multi-robot task allocation method for time-constrained dynamic collective transport,'' \emph{Rob. Auton. Syst.}, vol. 178, p. 104722, 2024.

\bibitem{wu2024proof}
E.~W. Wu \emph{et~al.}, ``Proof of location verification towards trustworthy collaborative multi-vendor robotic systems,'' in \emph{ICIT}.\hskip 1em plus 0.5em minus 0.4em\relax IEEE, 2024, pp. 1--8.

\bibitem{lamballais2017estimating}
T.~Lamballais, D.~Roy, and M.~De~Koster, ``Estimating performance in a robotic mobile fulfillment system,'' \emph{Eur. J. Oper. Res.}, vol. 256, no.~3, pp. 976--990, 2017.

\bibitem{hairoboticsHaiPickSystem}
``Hairobotics-automated case-handling mobile robot systems,'' \url{https://www.hairobotics.com/products/haipick-system-1}, [Accessed 15-02-2024].

\bibitem{qin2024performance}
Z.~Qin \emph{et~al.}, ``Performance analysis of multi-tote storage and retrieval autonomous mobile robot systems,'' \emph{Transp. Sci.}, 2024.

\bibitem{chi2021analysis}
C.~Chi \emph{et~al.}, ``Analysis and optimization of the robotic mobile fulfillment systems considering congestion,'' \emph{Appl. Sci.}, vol.~11, no.~21, p. 10446, 2021.

\bibitem{wang2022performance}
K.~Wang \emph{et~al.}, ``Performance evaluation of a robotic mobile fulfillment system with multiple picking stations under zoning policy,'' \emph{CAIE}, vol. 169, p. 108229, 2022.

\bibitem{yang2022modelling}
P.~Yang, G.~Jin, and G.~Duan, ``Modelling and analysis for multi-deep compact robotic mobile fulfilment system,'' \emph{Int. J. Prod. Res.}, vol.~60, no.~15, pp. 4727--4742, 2022.

\bibitem{wu2020research}
S.~Wu \emph{et~al.}, ``Research of the layout optimization in robotic mobile fulfillment systems,'' \emph{Int. J. Adv. Robot. Syst.}, vol.~17, no.~6, p. 1729881420978543, 2020.

\bibitem{zou2018evaluating}
B.~Zou \emph{et~al.}, ``Evaluating battery charging and swapping strategies in a robotic mobile fulfillment system,'' \emph{Eur. J. Oper. Res.}, vol. 267, no.~2, pp. 733--753, 2018.

\bibitem{roy2019robot}
D.~Roy \emph{et~al.}, ``Robot-storage zone assignment strategies in mobile fulfillment systems,'' \emph{Transp. Res. E Logist. Transp. Rev.}, vol. 122, pp. 119--142, 2019.

\bibitem{lamballais2020inventory}
T.~Lamballais~Tessensohn, D.~Roy, and R.~B. De~Koster, ``Inventory allocation in robotic mobile fulfillment systems,'' \emph{IISE Trans.}, vol.~52, no.~1, pp. 1--17, 2020.

\bibitem{duan2021performance}
G.~Duan \emph{et~al.}, ``Performance evaluation for robotic mobile fulfillment systems with time-varying arrivals,'' \emph{CAIE}, vol. 158, p. 107365, 2021.

\bibitem{bolch2006queueing}
G.~Bolch \emph{et~al.}, \emph{Queueing networks and Markov chains: modeling and performance evaluation with computer science applications}.\hskip 1em plus 0.5em minus 0.4em\relax John Wiley \& Sons, 2006.

\bibitem{metropolis1949monte}
N.~Metropolis and S.~Ulam, ``The monte carlo method,'' \emph{Journal of the American statistical association}, vol.~44, no. 247, pp. 335--341, 1949.

\bibitem{buitenhek2000amva}
R.~Buitenhek, G.-J. van Houtum, and H.~Zijm, ``Amva-based solution procedures for open queueing networks with population constraints,'' \emph{Ann. Oper. Res.}, vol.~93, no. 1-4, pp. 15--40, 2000.

\bibitem{AnyLogicSimulation}
``{A}ny{L}ogic: {S}imulation {M}odeling {S}oftware {T}ools \& {S}olutions for {B}usiness --- anylogic.com,'' \url{https://www.anylogic.com/}, [Accessed 28-02-2024].

\end{thebibliography}

\end{document}